\documentclass[runningheads]{llncs}

\usepackage[T1]{fontenc}
\usepackage{graphicx}
\usepackage{amsmath}
\usepackage{amsfonts}
\usepackage{float}
\usepackage{multirow}

\usepackage{color}
\usepackage[colorlinks=true, citecolor=blue,
            linkcolor=blue, urlcolor=blue]{hyperref}

\begin{document}

\title{Graph-XGB: A Hybrid GNN Framework for
Outdoor-to-Indoor 5G Signal Reconstruction via
Aerial-Only UAV RF Sensing}

\titlerunning{Graph-XGB for O2I Signal Reconstruction}

\author{David Hason Rudd\inst{1} \and
Eunkwang Shin\inst{1} \and
Nithiwat Pattrapong\inst{1} \and
Huan Huo\inst{1} \and
Xiangmeng Wang\inst{2}}

\authorrunning{D. Hason Rudd et al.}

\institute{University of Technology Sydney,
15 Broadway, Ultimo NSW 2007, Australia\\
\email{\{david.hasonrudd,huan.huo\}@uts.edu.au}\\
\email{\{eunkwang.shin,nithiwat.pattrapong\}@student.uts.edu.au}
\and
The Hong Kong Polytechnic University,
Hung Hom, Kowloon, Hong Kong SAR\\
\email{xiangmengpoly.wang@polyu.edu.hk}}

\maketitle
{\let\thefootnote\relax\footnotetext{This is the authors' accepted manuscript of a chapter published in BESC 2026, Lecture Notes in Computer Science, vol.~17175, pp.~18--37, Springer, 2027. It is not the Version of Record and does not reflect post-acceptance improvements or corrections. The Version of Record is available online at \url{https://doi.org/10.1007/978-981-92-6910-5_2}.}}

\begin{abstract}
Delivering robust In-Building Coverage (IBC) is critically impeded
by severe Outdoor-to-Indoor (O2I) attenuation, a constraint that
intensifies as deployments move from 5G mid-band towards mmWave and
prospective 6G sub-THz carriers, where losses of 60--80~dB have been
reported across modern building envelopes. While Unmanned Aerial
Vehicles (UAVs) offer scalable exterior RF sensing platforms, a
fundamental inversion problem persists: interior 3D signal states
cannot be directly sampled from constrained exterior flight paths.
This paper proposes Graph-XGB, a hybrid inference framework that
decouples spatial topology reasoning from non-linear signal
regression to resolve this challenge. A modified SpiderGNN encoder,
operating on a topology-aware graph constructed with a physically
grounded \emph{Floor-Aware 3D Distance} metric and Bridge Edge
Guarantees, propagates UAV boundary measurements into
128-dimensional indoor embeddings. Principal Component Analysis
(PCA) compresses these into a 10-dimensional manifold retaining
96.6\% of embedding variance, fused with raw spatial features and
decoded by independent, physics-tailored XGBoost regressors. This
decoupling mitigates the optimization conflict between smoothly
decaying path-loss signals and volatile interference-dependent
metrics. Evaluated on a real-world 5G NR dataset of 613 verified
nodes spanning the 10 floors of a single building, Graph-XGB attains
the strongest accuracy among the architectures compared, with $R^2$
of 0.759 (RSSI) and 0.770 (RSRP), improvements of 35.5\% and
64.9\% over the SimpleGNN baseline, using no interior RF
measurement as a model input; indoor observations enter only as
training supervision, with evaluation on held-out internal nodes.
Interference-dependent targets (RSRQ, SNR and CQI) remain bounded
well below the path-loss metrics, delimiting the reach of purely
spatial inference. The framework further enables continuous 3D
volumetric coverage maps for null-zone identification in complex
multi-story environments.

\end{abstract}

\section{Introduction}

With the deployment of fifth-generation (5G) cellular networks, the
telecommunications industry has shifted network planning from
macro-level horizontal coverage to highly granular,
three-dimensional (3D) spatial capacity. While historical heuristics
cited over 80\% of mobile data consumption as occurring indoors,
recent empirical analyses indicate a stabilization around 70--75\%
due to outdoor autonomous mobility and mobile Extended Reality (XR)
applications \cite{5gamericas2024neutral}; nevertheless, the
absolute volume and density of indoor demand have grown
exponentially \cite{ericsson2025mobility}. Delivering robust
In-Building Coverage (IBC) therefore remains a paramount hurdle for
Mobile Network Operators (MNOs), particularly in dense urban canyons
and multi-story high-rise structures where macro-cellular
architectures optimized for ground-level propagation struggle to
achieve effective vertical penetration.

The barrier is physical. Contemporary high-rise construction relies
on reinforced concrete cores, steel-framed facades and metal-coated,
low-emissivity (low-E) glass, which act as unintentional Faraday
cages inducing severe electromagnetic reflection and absorption.
Whereas 5G mid-band penetration losses typically range from 20 to
45 dB, 6G sub-Terahertz (sub-THz) signals frequently experience
attenuation exceeding 60 to 80 dB when traversing low-E glass and
concrete~\cite{xing2021sub}. Signals from exterior macro-cells
consequently degrade below the noise floor within a few meters of
the building envelope, producing severe ``signal space ambiguity''
and unpredictable dead zones within interior volumes.

Mapping this complex RF environment has traditionally required
labor-intensive walk-tests using specialized measurement equipment
\cite{infovista2025tems}: accurate, but strictly localized, highly
unscalable, and economically unfeasible for broad network
optimization or digital twin synchronization~\cite{ficili2025sensors}.
Unmanned Aerial Vehicles (UAVs) address this scalability deficit as
highly agile 3D RF sensing platforms~\cite{tarhuni2023machine},
executing granular multi-altitude scans of building exteriors beyond
the reach of static ground-based setups. They also impose a hard
physical constraint: UAVs are strictly confined to exterior flight
paths and cannot directly sample the internal building environment.
This precipitates the core ``O2I Inversion Problem'': can the
internal 3D signal state of a multi-story structure be accurately
reconstructed relying solely on sparse, exterior aerial boundary
measurements?

As detailed in Section~2, existing data-driven approaches do not
resolve this. Physically rigorous ray-tracing models remain
impractical for dynamic environments, while deep learning methods
predominantly target indoor-to-indoor interpolation over known
Access Point (AP) geometries and fixed Euclidean grids, and thus
lack the topological agility required for the asymmetric boundary
transitions of UAV-driven O2I penetration. Graph Neural Networks
(GNNs) abandon static grids in favour of attention-derived spatial
graphs, yet a critical bottleneck persists in end-to-end
architectures. Operators require predictions of both path-loss
dominated metrics (RSSI, RSRP) and interference-dependent quality
metrics (RSRQ, SNR, CQI), and our prior experiments revealed that
predicting these fundamentally divergent signal physics within a
monolithic network induces a severe optimization conflict: the model
cannot balance the smoothly decaying, distance-dependent nature of
RSRP against the highly volatile, load-dependent fluctuations of
CQI, resulting in a strict performance ceiling.

We therefore propose Graph-XGB, a hybrid inference framework for 3D
IBC prediction that deliberately decouples spatial topology
reasoning from non-linear signal regression. The framework is
developed and validated on 5G NR measurements and is motivated by
the intensifying O2I penetration challenge anticipated in 6G
deployments. A recursive Graph Attention Network operates
exclusively as an O2I propagation encoder: leveraging a newly
designed floor-aware 3D distance metric and bridge-edge guarantees,
it projects the UAV boundary signals deep into the building
manifold. These embeddings are compressed via Principal Component
Analysis (PCA), fused with raw spatial coordinates, and decoded by
independent, target-specific eXtreme Gradient Boosting (XGBoost)
regressors tuned to the physics of each target signal.

The primary contributions of this work are summarized as follows:
\begin{itemize}
  \item \emph{Decoupled Hybrid Architecture:} We shift from
    end-to-end GNN prediction to a hybrid GNN-XGBoost framework,
    utilizing the graph strictly as an O2I latent feature encoder
    and thereby resolving the optimization conflict between
    path-loss and interference targets.
  \item \emph{Physically Grounded Topology Construction:} We
    formulate a novel Floor-Aware 3D Distance metric that
    geometrically penalizes vertical floor separation to reflect
    structural attenuation, augmented by Bridge Edge Guarantees that
    prevent information starvation for deep internal nodes.
  \item \emph{Target-Specific Regression:} Projecting graph
    embeddings into a low-dimensional manifold via PCA enables
    independent, physics-tailored XGBoost ensembles, shallow trees
    for stable RSSI/RSRP prediction, deep trees for complex SNR/CQI
    extraction.
  \item \emph{UAV-to-Indoor Empirical Validation:} We validate on a
    multi-modal dataset of real-world UAV exterior drone flights and
    internal TEMS walk-tests from a single 10-floor building. The
    measurements are 5G NR, treated as a physically representative
    proxy for the 6G IBC planning challenge rather than as direct
    sub-THz evidence. The framework predicts all five signal
    parameters at held-out internal nodes using no interior RF
    measurement as an input, with high accuracy on the path-loss
    targets while the interference-dependent targets remain bounded
    by a spatial-inference ceiling we characterise empirically.
\end{itemize}

\section{Related Works}

Delivering consistent In-Building Coverage (IBC) remains a
formidable technical hurdle, specifically within dense urban canyons
and vertical high-rise
structures~\cite{5gamericas2024neutral,wang2014cellular}.
Traditional macro-cellular architectures, optimized for horizontal
propagation, struggle to achieve effective vertical penetration into
multi-story edifices~\cite{shakya2022dense,karner2006indoor}. The
architectural complexity of multi-floor environments introduces
severe signal space ambiguity, exposing the limitations of standard
2D spatial models when tracking vertical signal
degradation~\cite{mostafa2025survey}. This challenge is compounded
by modern construction materials designed for thermal and acoustic
insulation~\cite{ng2022electromagnetic,ayeni2022correlation,%
pavlik2024influence}. Elements such as reinforced concrete,
metal-coated glass, and steel frames induce severe attenuation,
frequently dampening signals by 20 to 45 dB~\cite{asp2012radio}.
Recent drone-based empirical measurements utilizing 5G mmWave
frequencies further demonstrate that Outdoor-to-Indoor (O2I)
signals can drop below the noise floor within just 10 meters of
indoor penetration~\cite{fuschini2022multiband}. Furthermore,
issues like pilot contamination and inter-cell interference
complicate the signal landscape in elevated
spaces~\cite{ghanem2012reducing}.

While solutions like Distributed Antenna Systems (DAS) and
femtocells exist, they are often characterized by high capital
expenditure and structural rigidity, limiting their
scalability~\cite{ma2024reconfigurable,jafari2015small,%
paulo2021study}. Conversely, Unmanned Aerial Vehicles (UAVs) have
been identified as agile platforms for RF
sensing~\cite{fotouhi2019survey}. UAVs equipped with measuring
apparatus can perform granular 3D scanning of building
exteriors~\cite{martian2024direction}, overcoming the height and
placement limitations of traditional ground-based setups to capture
true 3D coverage profiles.

However, a key limitation remains: UAVs are restricted to outdoor flight paths and are unable to directly measure the signal conditions inside buildings. As a result, existing approaches still depend heavily on labor-intensive walk tests to characterize indoor coverage~\cite{rocha2023ziwi}. This creates a fundamental inversion problem: Can the internal signal state be accurately reconstructed
solely from sparse external aerial measurements? Recent groundbreaking work in O2I bridging has demonstrated that exterior signal measurements can successfully reconstruct 3D indoor WiFi Access Point configurations without internal reference
points~\cite{amano2025referencefree}. However, while successful in physical localization tasks, inferring complex, interference-dependent coverage metrics (such as CQI and RSSI distribution) across entire 3D multi-floor topologies requires advanced spatial reasoning.

Recent efforts utilizing deep learning have aimed to bridge the gap in various fields by enhancing the accuracy and efficiency of data analysis and pattern recognition~\cite{kurner2002prediction,maduraga2021comparison,%
raj2021indoor,konak2024wireless}. To better capture complex
propagation paths, recent advancements have explored generative
architectures, ranging from standard diffusion models for indoor
map interpolation (IRDM)~\cite{qiu2023irdm} to physics-informed
networks that explicitly encode building material transmission
coefficients and multi-path priors, guided by the Helmholtz
equation (iRadioDiff,
RadioDiff-k2)~\cite{wang2025iradiodiff,wang2026radiodiff}. However,
these generative methods predominantly focus on indoor-only modeling
based on known Access Point (AP) geometries and fixed Euclidean
grids. Consequently, they lack the topological agility required to
solve the O2I inversion problem from sparse, exterior aerial
measurements. More broadly, the radio-map estimation and
coverage-prediction literature, empirical path-loss regressions,
geostatistical interpolators such as kriging, and learned radio-map
construction, assumes that the sampling support intersects the
prediction domain. The O2I inversion setting violates that
assumption by construction, since the entire interior volume lies
outside the sampled support; we therefore reach interior states by
multi-hop propagation from an exterior boundary rather than by
interpolation within a sampled field.

Conversely, Graph Neural Networks (GNNs) have demonstrated superior
capability in modeling non-Euclidean spatial relationships in indoor
positioning tasks~\cite{zhang2023domain,lezama2021indoor,%
lezama2023application,vishwakarma2023indoorgnn}. Early GNN
applications often relied on static, predefined graph structures
that struggle to adapt to the highly variable physical constraints
of complex indoor environments. To address this, recent innovations
have leveraged dynamic graph generation, such as processing inputs
in learned lower-dimensional spaces
(GravNetConv)~\cite{bayraktar2024indoor} and integrating
physics-informed latency architectures
(GraphKANLoc)~\cite{zhang2026graphkanloc}, to capture more
fine-grained spatial dependencies.

\begin{figure}[ht!]
\centering
\includegraphics[width=\linewidth]{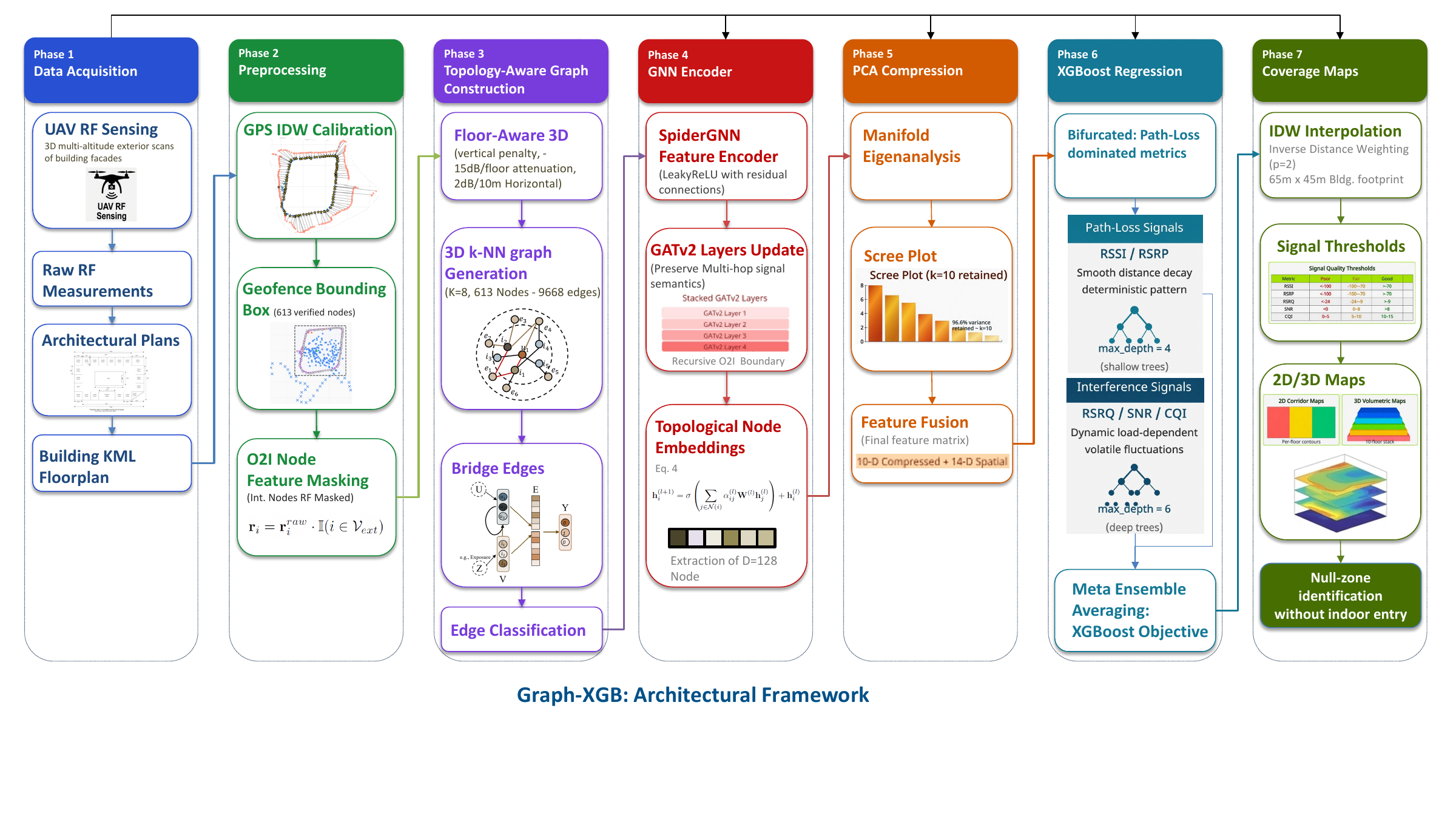}
\caption{The Graph-XGB Architectural Framework Overview.
The pipeline integrates multi-modal data acquisition with latent
topology inference to guide a recursive GAT encoder. To prevent
optimization conflict, the framework introduces a bifurcated
decoding stage utilizing decoupled XGBoost regressors that
independently reconstruct, in parallel, the path-loss (RSSI, RSRP)
and interference-dependent (RSRQ, SNR, CQI) coverage maps from the
compressed topological subspace. \emph{Note: The diagram also illustrates the cascaded dependency flow, an experimental variant (v9) explored alongside our primary independent architecture (v8).}}
\label{fig:overview}
\end{figure}

Building upon these topological foundations, recent models have
introduced dynamic edge reasoning (SpiderGNN)~\cite{gao2025spidergnn}
to build sparse structures dynamically using attention-derived
feature affinity. In this work, we advance this trajectory by
proposing a comprehensive seven-phase inference framework leveraging a
Dynamic Graph Attention Network (DGAT)~\cite{rudd2025location}. By
abandoning static geometric assumptions in favor of learned
topological associations, our model captures the latent propagation
dynamics required to holistically infer 3D indoor coverage from the
outside. A high-level overview of the proposed Graph-XGB
architecture is illustrated in Figure~\ref{fig:overview}.

\section{Methodology}

To overcome the limitations inherent in purely geometric spatial
models and end-to-end deep learning approximations, we propose the
Graph-XGB inference framework. This architecture deliberately
decouples topological feature extraction from non-linear signal
regression. Formally, we reconceptualize the Graph Attention Network
(GAT) not as a direct predictor, but as an Outdoor-to-Indoor (O2I)
propagation encoder $f_{\theta}$. The resulting topology-aware
embeddings are then dimensionally reduced via a projection function
$g_{\phi}$ and processed by independent, target-specific
gradient-boosted trees $h_{\psi}$.

\subsection{Data Acquisition, Calibration, and O2I Asymmetry}

The foundation of our dataset comprises external drone-based Radio
Frequency (RF) scans and corresponding internal walk-test
validations, predicting five continuous signal quality targets
(RSSI, RSRP, RSRQ, SNR, and CQI) alongside architectural KML
floorplans.

Raw measurement logs frequently suffer from severe indoor GPS
multipath drift and sensor artifacts. To establish a physically
rigorous foundation, the spatial coordinates underwent an Inverse
Distance Weighting (IDW) calibration process using a surveyed 3D
reference path. Subsequently, standard physics-based consistency
filters (derived from 3GPP TS~36.133) and geofencing bounding boxes
were applied to strip impossible signal-to-noise anomalies. To
ensure complete reproducibility, the full data curation pipeline,
reverse-engineering parsing scripts, and the raw datasets are
open-sourced and publicly available\footnote{Code and datasets are
available at the following GitHub repository:
\url{https://github.com/uts-aiot-ibc-research/graph-xgb}}.

Let the curated spatial environment be modeled as a directed graph
$\mathcal{G} = (\mathcal{V}, \mathcal{E})$, where the node set
$\mathcal{V}$ is strictly partitioned into external observation
nodes $\mathcal{V}_{ext}$ and internal target nodes
$\mathcal{V}_{int}$. To operationalize the O2I learning objective,
a strict asymmetric feature masking strategy is employed. Let the
feature matrix for node $i$ be $x_i = [s_i \parallel r_i]$,
representing the concatenation of spatial parameters $s_i$ and the
RF signal vector $r_i$. We apply an indicator masking function to
the RF vector:
\begin{equation}
  r_i = r_i^{raw} \cdot \mathbb{I}(i \in \mathcal{V}_{ext})
\end{equation}
where $\mathbb{I}(\cdot)$ is the indicator function. Consequently,
for all $u \in \mathcal{V}_{int}$, $r_u = 0$, forcing the
downstream network to infer indoor states entirely from external
boundary propagation and strictly preventing interior RF values from
entering the model as input features. This masking governs the
\emph{inputs} only: a disjoint training subset of internal nodes
supplies target values as supervision, so the reported results
characterise reconstruction at held-out interior locations within an
instrumented building rather than transfer to a building for which
no interior measurement exists.

\subsection{Topology-Aware Graph Construction}

Standard $k$-Nearest Neighbor ($k$-NN) algorithms utilizing 2D
Haversine or 3D Euclidean metrics fail to capture the asymmetric
attenuation of multi-story structures. Vertical propagation through
reinforced concrete floors introduces significantly higher path loss
(approximately 15 dB per floor) compared to horizontal propagation
(approximately 2 dB per 10 meters).

To physically ground the graph topology, we define a \emph{Floor-Aware
3D Distance} metric for 3D $k$-NN edge generation ($k=8$). The
adjusted distance $d_{\text{3D}}$ between nodes $i$ and $j$ is
formulated as:
\begin{equation}
  d_{\text{3D}}(i, j) =
    \sqrt{d_{\text{hav}}(i,j)^2 + (\eta \cdot \Delta f_{ij})^2}
\end{equation}
where $d_{\text{hav}}(i,j)$ is the geographic Haversine distance,
$\Delta f_{ij} = |f_i - f_j|$ represents the discrete floor
separation, and $\eta = 75$~meters is an empirical vertical penalty
factor derived from the established signal physics, equating the
15~dB vertical attenuation of a single floor to 75~meters of
horizontal traversal (at 2~dB per 10~meters).

Relying strictly on standard $k$-NN causes structural isolation for
deep internal nodes, cutting them off from the external RF boundary.
Let $\mathcal{N}_{ext}(u) = \{v \in \mathcal{V}_{ext} \mid e_{uv}
\in \mathcal{E}\}$ denote the set of external neighbors for an
internal node $u$. To prevent information starvation, we enforce a
\emph{Bridge Edge Guarantee}:
\begin{equation}
  \text{If } \mathcal{N}_{ext}(u) = \emptyset,\;
  \text{append } e_{uv^*} \text{ to } \mathcal{E}
  \text{ where }
  v^* = \underset{v \in \mathcal{V}_{ext}}{\text{argmin}}\;
  d_{\text{3D}}(u, v)
\end{equation}

Finally, every directed edge $e_{ij}$ is assigned a categorical
classification $\tau_{ij} \in \{0, 1, 2, 3\}$ denoting its
propagation context (ext-to-ext, ext-to-int, int-to-ext,
int-to-int). The resulting edge attribute vector is defined as
$e_{ij} = [d_{\text{3D}}(i,j),\; \Delta f_{ij},\; \tau_{ij}]^\top$.

\subsection{The SpiderGNN Feature Encoder}

We deploy a modified SpiderGNN architecture utilizing stacked Graph
Attention Network v2 (GATv2) layers. Operating exclusively as a
spatial feature extractor, the network recursively aggregates
boundary signals across the floor-aware propagation manifold.

At layer $l$, the node representations $h_i^{(l)}$ are updated
using dynamic attention mechanisms that incorporate both node
features and edge attributes to preserve multi-hop signal
semantics:
\begin{equation}
  h_i^{(l+1)} = \sigma\!\left(
    \sum_{j \in \mathcal{N}(i)} \alpha_{ij}^{(l)}\, W^{(l)}\,
    h_j^{(l)}
  \right) + h_i^{(l)}
\end{equation}
where $\sigma$ is a non-linear LeakyReLU activation and the
attention coefficients $\alpha_{ij}$ are computed via a scoring
function on the concatenated feature space. The residual term
$+h_i^{(l)}$ mitigates over-smoothing. The final output is a dense,
high-dimensional node embedding matrix $H \in \mathbb{R}^{N \times
D}$ (where $D=128$), encapsulating the complex O2I propagation
context for the entire building manifold.

\subsection{Dimensionality Reduction via PCA and Feature Fusion}

Eigen-analysis of the covariance matrix of $H$ reveals that the O2I
propagation problem occupies a strict low-dimensional manifold.
Principal Component Analysis (PCA) scree plots demonstrate that the
effective rank of the embedding space is small, with the first
$k=10$ principal components accounting for approximately 96.6\% of
the total variance.

We formalize the dimensionality reduction by computing the centered
covariance matrix $C = \frac{1}{N}\bar{H}^\top \bar{H}$ and its
eigendecomposition $C = V\Lambda V^\top$. The raw embeddings are
then projected into a compressed subspace:
\begin{equation}
  H_{PCA} = \bar{H}\, V_k
\end{equation}
where $V_k \in \mathbb{R}^{D \times 10}$ contains the top 10
eigenvectors. This step isolates the dominant axes of signal
propagation, stripping high-dimensional structural noise that
disrupts downstream decision-tree regressors.

To construct the final feature space for downstream regression,
these compressed topological embeddings are concatenated with the
original raw node features (retained as a spatial baseline safeguard). Let $X_{raw} \in \mathbb{R}^{N \times
14}$ denote the matrix of raw spatial and structural parameters
(e.g., latitude, longitude, and floor index). The final composite
feature matrix $Z \in \mathbb{R}^{N \times 24}$, which serves as
the direct input to the XGBoost models, is defined as:
\begin{equation}
  Z = [H_{PCA} \parallel X_{raw}]
\end{equation}
This feature fusion ensures that the regressors benefit
simultaneously from the non-Euclidean propagation context extracted
by the GNN and the absolute geometric precision of the original
spatial coordinates.

\subsection{Target-Specific XGBoost Regression \& Ensembling}

In the final bifurcated stage, the conventional GNN multi-layer perceptron heads are replaced with independent XGBoost regressors $h_{\psi}$, applied to the fused feature matrix $Z$. The objective function to be minimized for each target $y$ integrates a convex loss function $l$ and a regularization term $\Omega$:
\begin{equation}
  \mathcal{L}(\psi) =
    \sum_{i=1}^{N} l(y_i, \hat{y}_i) +
    \sum_{m=1}^{M} \Omega(f_m)
\end{equation}
where $\hat{y}_i = \sum_{m=1}^{M} f_m(z_i)$ is the prediction from
an ensemble of $M$ trees. This hybrid design enables critical
target-specific decoupling:
\begin{itemize}
  \item \emph{Path-Loss Dominated Signals
    ($\mathcal{Y}_{path} \in \{\text{RSSI, RSRP}\}$):}
    Varying smoothly with distance decay and deterministic structural
    penetration, these metrics utilize conservative shallow trees
    (max\_depth~=~4) to prevent overfitting on local spatial
    anomalies.
  \item \emph{Interference-Dependent Signals
    ($\mathcal{Y}_{interf} \in \{\text{RSRQ, SNR, CQI}\}$):}
    Highly volatile and load-dependent due to neighbor cell
    interference, these metrics utilize deep tree topologies
    (max\_depth~=~6) to extract localized multi-path interactions.
\end{itemize}

To suppress stochastic variance inherent in GNN initialization, the
decoupled pipeline is executed within a meta-ensemble
framework—designated as the proposed Graph-XGB model (v8). The final prediction $\hat{y}_{final}$ is the uniform
average over $S=9$ independent initializations:
\begin{equation}
  \hat{y}_{final} = \frac{1}{S}\sum_{s=1}^{S} \hat{y}^{(s)}
\end{equation}

Furthermore, to explore the Shannon-Hartley relationship, we
extend this architecture into an experimental cascaded variant (v9),
where RSRP predictions are explicitly fed as subsequent input features
into the interference-dependent regressors (RSRQ, SNR, and CQI).

\section{Experimental Setup}

\subsection{Dataset Description \& Stratified Splitting}

\begin{figure}[!h]
\centering
\includegraphics[width=0.8\textwidth]{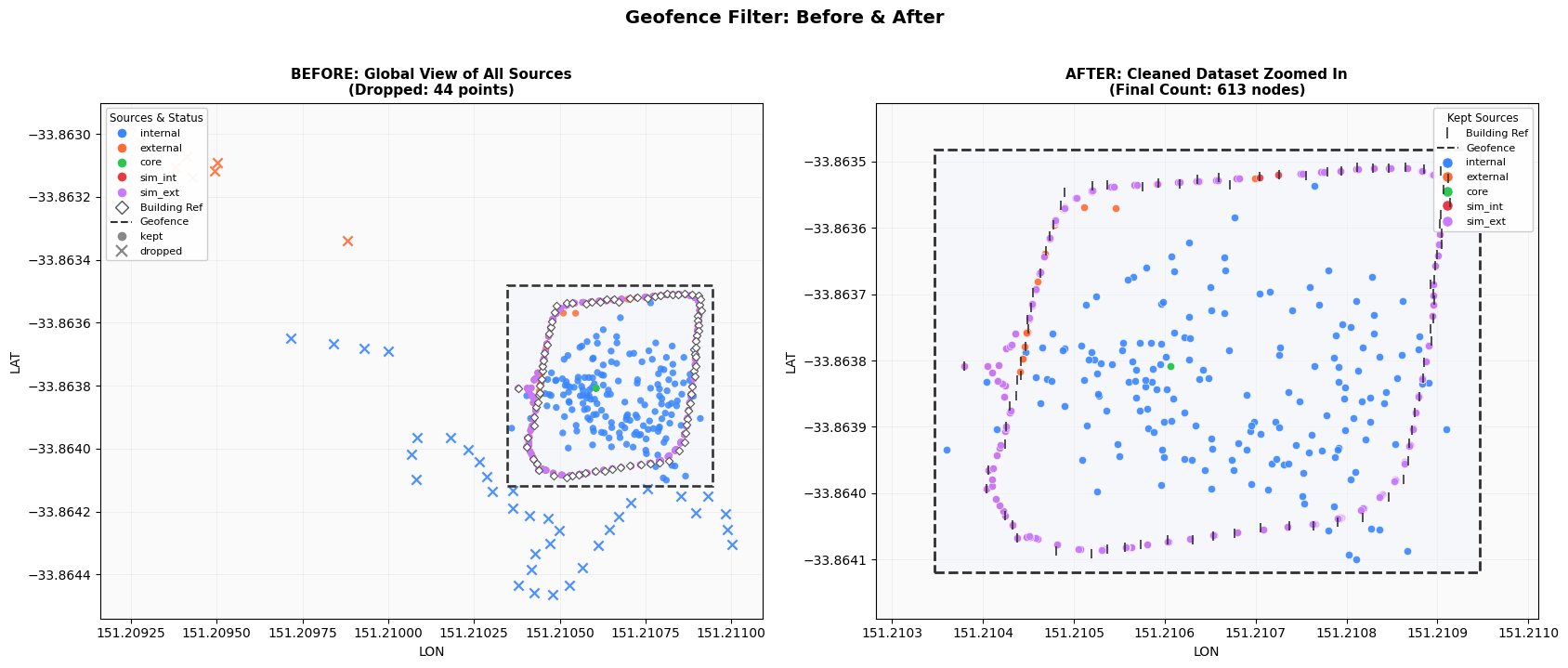}
\caption{Geofence Filtering: Left panel shows the raw data
distribution with significant GPS drift; right panel displays the
cleaned dataset (613 nodes) after applying the structural bounding
box (dashed line) and physical consistency
filters.}\label{fig:geofence}
\end{figure}

The finalized spatial graph comprises 613 verified nodes and 9,668
directed edges. The dataset construction involved a rigorous
cleaning pipeline (illustrated in Fig.~\ref{fig:geofence}) to
mitigate indoor GPS multipath drift and eliminate sensor artifacts.

\begin{figure}[!h]
\centering
\includegraphics[width=0.85\textwidth]{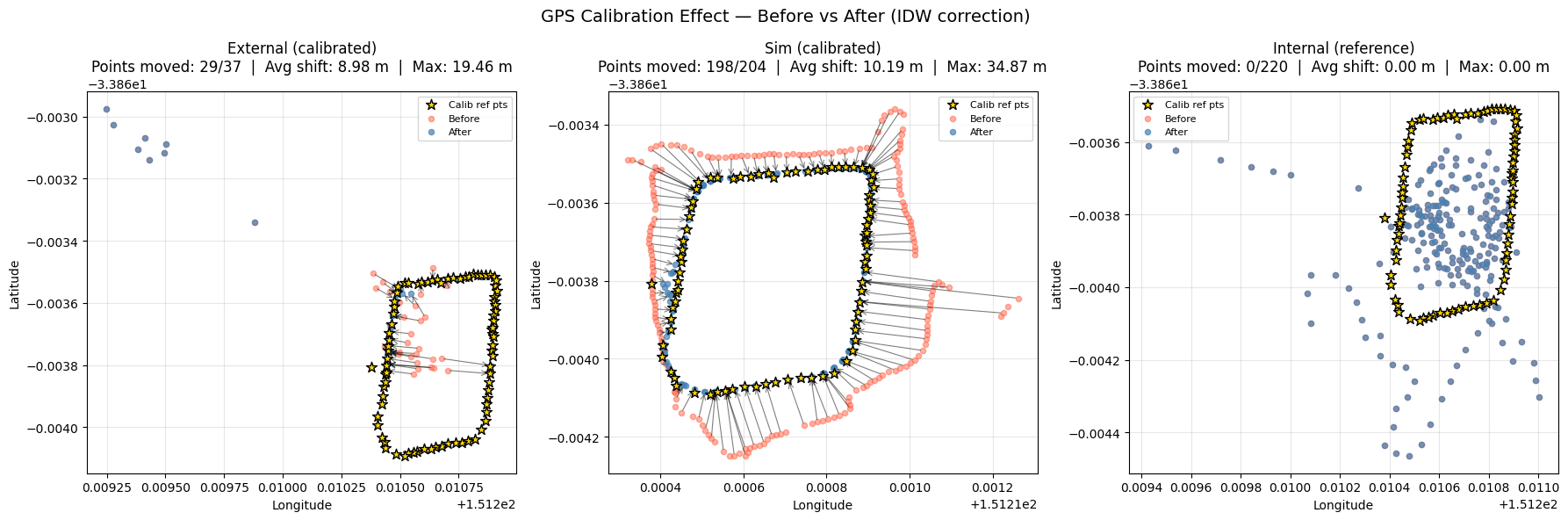}
\caption{GPS Calibration Effect: The IDW correction redistributes
drifted measurement points toward the surveyed reference path
(stars), compensating for multipath errors across the external and
internal measurement sets.}\label{fig:gps_calib}
\end{figure}

The spatial coordinates underwent Inverse Distance Weighting (IDW)
calibration to correct severe indoor drift by realigning
measurements to known physical reference paths
(Fig.~\ref{fig:gps_calib}).

To ensure robust model evaluation and address statistical
reliability within a constrained dataset, the 613 nodes were
strictly partitioned by node type. The 229 external nodes served
exclusively as boundary signal providers (with masked targets during
training). The remaining 384 internal target nodes were evaluated.
To prevent unbalanced RF distributions, the 384 internal nodes were
partitioned into training (70\%, 268 nodes), validation (15\%,
58 nodes), and test (15\%, 58 nodes) sets using a stratified split
on the first principal component of the joint RF target space
(PCA-derived stratification variable) to ensure balanced
signal-strength distributions across all three partitions. All
partitions derive from the same building and measurement campaign;
the protocol therefore evaluates reconstruction at unobserved
interior locations within an instrumented structure, not
cross-building transfer.

\subsection{Baseline Models and Architectural Ablation}
To rigorously evaluate the proposed architecture, we benchmark
performance against established paradigms and conduct an internal
architectural ablation study to validate the hybrid framework:
\begin{itemize}
  \item \textbf{SimpleGNN (GraphSAGE):} A 2-layer GraphSAGE baseline
    utilizing only graph topology without edge type features. This
    establishes the performance floor for graph-based inference.
  \item \textbf{End-to-End SpiderGNN:} The direct predecessor to our
    hybrid model. It utilizes GATv2 layers directly connected to an
    MLP regression head, trained end-to-end without feature
    decoupling or PCA compression.
  \item \textbf{Two-Stage Frozen Trunk (v3):} An ablation experiment
    where the SpiderGNN backbone is trained, its weights frozen, and
    independent MLP regression heads are subsequently trained on the
    embeddings.
  \item \textbf{Hybrid GNN-XGBoost PCA (v7):} Replaces the
    conventional MLP heads entirely. The network extracts raw
    embeddings, compresses them via PCA ($k=10$), and applies
    decoupled XGBoost regression.
  \item \textbf{Proposed Graph-XGB (v8):} Extends the v7 hybrid
    by applying target-specific XGBoost hyperparameters (shallow
    vs.\ deep trees) and uniformly averages predictions across 9
    independent neural initializations.
  \item \textbf{Cascaded XGBoost (v9):} Explores the Shannon-Hartley
    relationship by explicitly feeding the path-loss predictions
    (RSRP) as input features to the subsequent interference
    regressor models (RSRQ, SNR, and CQI).
\end{itemize}

\subsection{Implementation Details}

The proposed hybrid framework was implemented using PyTorch
Geometric for spatial graph extraction and the \textit{scikit-learn}
ecosystem for PCA and extreme gradient boosting. The spatial graph
was constructed using $k=8$ nearest neighbors, utilizing the
floor-aware 3D distance metric alongside strict bridge edge
guarantees to prevent internal node information starvation.

Crucially, the downstream XGBoost regressors are decoupled and
specifically tailored to the target physics. For path-loss dominated
targets (RSSI, RSRP), the tree depth was strictly constrained
(max\_depth~=~4) to prevent spatial overfitting on smooth signal
gradients. Conversely, for interference-dependent targets (RSRQ,
SNR, CQI), a deeper topology (max\_depth~=~6,
min\_child\_weight~=~1) was permitted to extract maximum non-linear
interactions before early stopping.

\subsection{Evaluation Metrics}

The predictive accuracy of the continuous regression targets was
evaluated using three standard metrics: Mean Absolute Error (MAE),
Root Mean Square Error (RMSE), and the Coefficient of Determination
($R^2$). They are mathematically defined as:
\begin{equation}
  \text{MAE} = \frac{1}{N}\sum_{i=1}^{N}\left|y_i - \hat{y}_i\right|
\end{equation}
\begin{equation}
  \text{RMSE} = \sqrt{\frac{1}{N}\sum_{i=1}^{N}
    \left(y_i - \hat{y}_i\right)^2}
\end{equation}
\begin{equation}
  R^2 = 1 -
    \frac{\sum_{i=1}^{N}\left(y_i - \hat{y}_i\right)^2}
         {\sum_{i=1}^{N}\left(y_i - \bar{y}\right)^2}
\end{equation}
where $y_i$ is the ground truth measurement, $\hat{y}_i$ is the
model prediction, and $\bar{y}$ is the mean of the observed data.

\section{Results and Discussion}

\subsection{Overall Predictive Performance and Architectural
Ablation}

The $R^2$ performance of the proposed hybrid architecture against
all baseline and ablation models is detailed in
Table~\ref{tab:overall_performance}, evaluated on the 58-node test
set. The comprehensive absolute error metrics (MAE and RMSE) for
all models across all five signal targets are reported in
Table~\ref{tab:mae_rmse}.

\begin{table}[ht!]
\caption{Overall Performance Comparison ($R^2$ Scores) across 5G
Signal Targets (58-node test set, single building). Best per column
in \textbf{bold}; ties are bolded
jointly.}\label{tab:overall_performance}
\centering
\begin{tabular}{|l|c|c|c|c|c|}
\hline
\textbf{Model Architecture}
  & \textbf{RSSI} & \textbf{RSRP}
  & \textbf{RSRQ} & \textbf{SNR} & \textbf{CQI} \\
\hline
SimpleGNN (GraphSAGE)
  & 0.560 & 0.467 & 0.181 & 0.125 & 0.079 \\
End-to-End SpiderGNN
  & 0.443 & 0.381 & 0.146 & 0.066 & 0.117 \\
Two-Stage Frozen Trunk (v3)
  & 0.622 & 0.609 & 0.232 & 0.042 & 0.093 \\
Hybrid GNN-XGBoost PCA (v7)
  & 0.732 & \textbf{0.775} & 0.230 & 0.090 & 0.164 \\
Cascaded XGBoost (v9)$^{\dagger}$
  & 0.746 & 0.763 & 0.262 & 0.121 & \textbf{0.195} \\
  \textbf{Proposed Graph-XGB (v8)}
  & \textbf{0.759} & 0.770
  & \textbf{0.268} & \textbf{0.146} & \textbf{0.195} \\
\hline
\end{tabular}
\vspace{1ex}
\raggedright\footnotesize

$^{\dagger}$v9 deliberately cascades RSRP predictions into
downstream regressors; the modest reduction in RSSI/RSRP performance
reflects upstream error propagation rather than architectural
failure, confirming v8 as the primary proposed model.
\end{table}

\begin{table}[ht!]
\caption{Comprehensive Error Metrics: MAE and RMSE across all
Architectural Variants (58-node test set, single building). RSSI and
RSRP in dBm; RSRQ and SNR in dB; CQI is dimensionless. Best per
column in \textbf{bold}; ties are bolded
jointly.}\label{tab:mae_rmse}
\centering
\footnotesize
\begin{tabular}{|l|c|c|c|c|c|c|}
\hline
\textbf{Model} & \textbf{Metric}
  & \textbf{RSSI} & \textbf{RSRP}
  & \textbf{RSRQ} & \textbf{SNR} & \textbf{CQI} \\
\hline
\multirow{2}{*}{SimpleGNN (GraphSAGE)}
  & MAE  & 6.81 & 7.34 & 1.39 & \textbf{4.65} & 2.57 \\
  & RMSE & 8.68 & 9.69 & 1.80 & 6.13 & 3.05 \\
\hline
\multirow{2}{*}{End-to-End SpiderGNN}
  & MAE  & 7.84 & 7.88 & 1.38 & 4.76 & 2.53 \\
  & RMSE & 9.76 & 10.44 & 1.84 & 6.33 & 2.99 \\
\hline
\multirow{2}{*}{Two-Stage Frozen Trunk (v3)}
  & MAE  & 6.19 & 6.04 & 1.33 & \textbf{4.65} & 2.50 \\
  & RMSE & 8.05 & 8.30 & 1.74 & 6.41 & 3.03 \\
\hline
\multirow{2}{*}{Hybrid GNN-XGBoost PCA (v7)}
  & MAE  & 5.27 & 4.67 & 1.31 & 4.86 & 2.36 \\
  & RMSE & 6.77 & \textbf{6.30} & 1.75 & 6.25 & 2.91 \\
\hline
\multirow{2}{*}{Cascaded XGBoost (v9)}
  & MAE  & \textbf{5.10} & 4.89
        & \textbf{1.25} & 4.76 & \textbf{2.33} \\
  & RMSE & 6.60 & 6.46 & 1.71 & 6.14 & \textbf{2.86} \\
\hline
\multirow{2}{*}{\textbf{Proposed Graph-XGB (v8)}}
  & MAE  & 5.16 & \textbf{4.62} & 1.27 & 4.76 & \textbf{2.33} \\
  & RMSE & \textbf{6.43} & 6.36
        & \textbf{1.70} & \textbf{6.06} & \textbf{2.86} \\
\hline
\end{tabular}
\end{table}

The empirical results systematically confirm that the proposed
GNN-XGBoost hybrid decisively outperforms all direct-regression GNN
baselines. A significant experimental finding from the ablation
study is that the core value of the GNN lies strictly in its ability
to aggregate the outdoor RF boundary signal through the spatial
graph into a compact embedding, rather than in its internal MLP
regression heads.

By replacing the neural regression heads with XGBoost ensembles
trained on PCA-reduced embeddings (v7), the framework immediately
surged to the highest $R^2$ scores observed across our ablation,
bypassing the overfitting
that caused early stopping in the end-to-end models. The
implementation of target-specific hyperparameters and the 9-seed
meta-ensemble (v8) named Graph-XGB further stabilised these results, achieving $R^2$
of 0.759 and 0.770 for RSSI and RSRP respectively, alongside the lowest RSSI RMSE (6.43 dBm) and a competitive RSRP RMSE (6.36 dBm, second only to v7's 6.30 dBm). As detailed in Table~\ref{tab:mae_rmse}, v8 achieves the
best RMSE in four of five signal targets and the best $R^2$ in four
of five targets (Table~\ref{tab:overall_performance}). Across the
nine initialisations the standard deviation of test $R^2$ is
$\pm0.016$ for path-loss metrics against $\pm0.089$ for the far more
volatile interference metrics; these quantify initialisation
variance alone, not confidence intervals over the 58-node test
sample. Relative to the SimpleGNN
baseline, v8 achieves a 35.5\% improvement in RSSI $R^2$
(0.759 vs.\ 0.560) and a 64.9\% improvement in RSRP $R^2$
(0.770 vs.\ 0.467). Against the end-to-end SpiderGNN predecessor,
the improvements reach 71\% for RSSI and exceed 100\% for RSRP,
confirming the decisive impact of the decoupled hybrid architecture.

When evaluating the Cascaded XGBoost architecture (v9), which
explicitly fed RSRP predictions as input features to the RSRQ, SNR,
and CQI regressors, we observed a nuanced trade-off. The cascade
introduces a modest upstream error propagation effect, marginally
reducing RSSI ($R^2 = 0.746$) and RSRP ($R^2 = 0.763$) relative to
the isolated v8 ensemble (differences of 0.013 and 0.007
respectively). Crucially, v9 fails to improve the variance capture ($R^2$) 
or RMSE for RSRQ and SNR over v8 (though it marginally edges v8 in RSRQ MAE), 
as reflected in Tables \ref{tab:overall_performance} and \ref{tab:mae_rmse}. 
The two architectures achieve identical CQI
performance ($R^2 = 0.195$, MAE = 2.33, RMSE = 2.86), confirming
that the Shannon-Hartley cascade provides negligible benefit under
purely spatial inference conditions. These findings identify
Graph-XGB (v8) as the strongest architecture among those evaluated,
leading on four of five targets (RSRP excepted, where v7 peaks).
This ranking rests on a single building and measurement campaign,
and the margin is decisive only for path-loss: for RSRQ, SNR and CQI
all variants remain within a narrow band of low $R^2$, so that
ordering is indicative rather than conclusive.

\begin{figure}[ht!]
\centering
\includegraphics[width=0.9\textwidth]{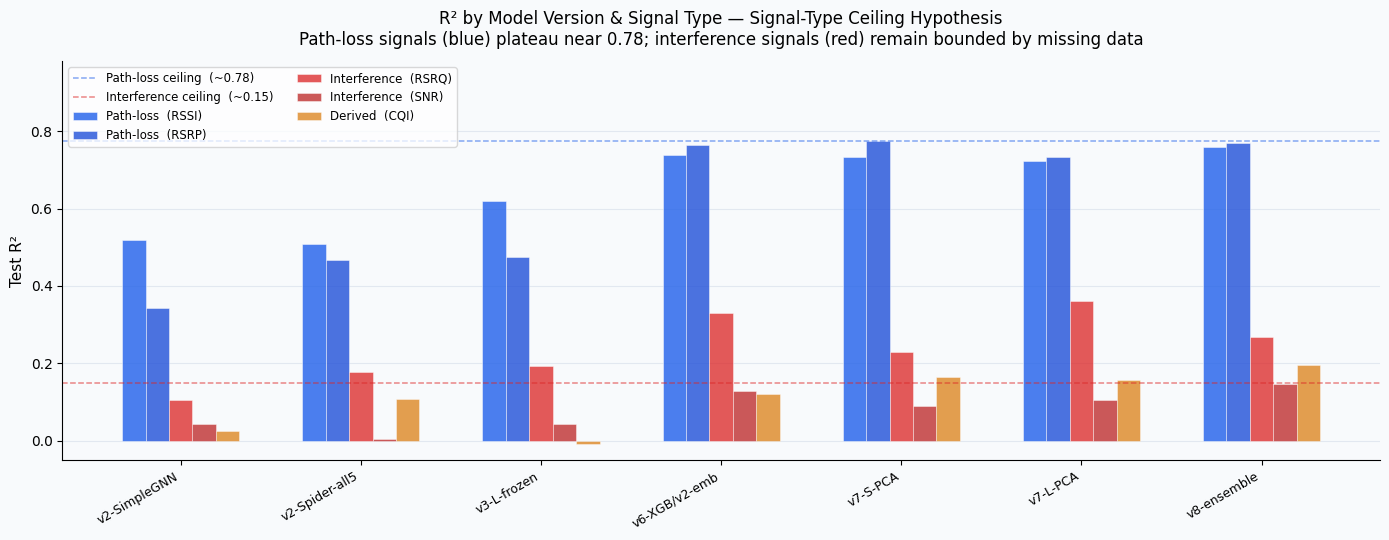}
\caption{$R^2$ evolution across model iterations. Path-loss signals
(RSSI, RSRP) climb from 0.44--0.56 to $\approx$0.77 at the v8
Graph-XGB, reaching a topological plateau (peak RSRP at v7).
Interference-dependent signals remain bounded, confirming the
Signal-Type Ceiling Hypothesis.}\label{fig:r2_ceiling}
\end{figure}

Tracking performance across architectural iterations
(Fig.~\ref{fig:r2_ceiling}) explicitly illustrates these divergent
learning trajectories. It establishes the efficacy of the
Outdoor-to-Indoor (O2I) spatial mapping framework for deterministic
power signals, while validating the physical limits of spatial-only
interference prediction (the Signal-Type Ceiling Hypothesis).

\begin{figure}[ht!]
\centering
\includegraphics[width=0.8\textwidth]{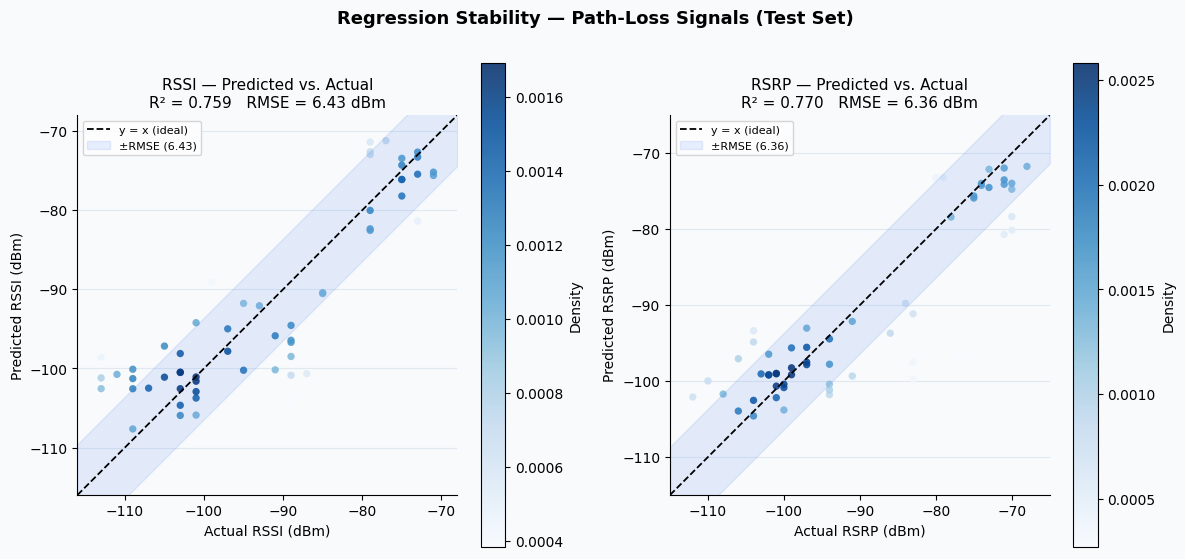}
\caption{Regression stability of the v8 Graph-XGB (58-node test
set). RSSI ($R^2=0.759$, RMSE~=~6.43~dBm) and RSRP ($R^2=0.770$,
RMSE~=~6.36~dBm) track the $y=x$ line over the core operational
range; the shaded band denotes $\pm$RMSE. Residuals are symmetric
about the diagonal, indicating no systematic
bias.}\label{fig:regression_stability}
\end{figure}

To further validate the physical reliability of these path-loss
predictions, we analysed the regression stability on the unseen test
set (Fig.~\ref{fig:regression_stability}). While $R^2$ confirms
variance capture, absolute error distribution is critical for
network planning. The scatter distributions demonstrate that both
RSSI and RSRP predictions cleanly track the $y=x$ identity line
across the core operational range in which the majority of test
nodes lie. Crucially, the residual spread is symmetrically
distributed above and below the diagonal, indicating that the
decoupled XGBoost regressor exhibits no systematic bias on this test
set, avoiding the over-prediction of weak signals or
under-prediction of strong signals that frequently plagues
end-to-end spatial networks.

\subsection{Latent Feature Extraction and the PCA Effective Rank}

A paramount insight derived from our architectural ablation is that
the primary value of the Graph Neural Network in O2I prediction lies
in its latent embedding quality. The SpiderGNN layers effectively
aggregate outdoor RF measurements through the spatial graph,
encoding the complex O2I propagation context into a dense
128-dimensional representation.

\begin{figure}[ht!]
\centering
\includegraphics[width=0.75\textwidth]{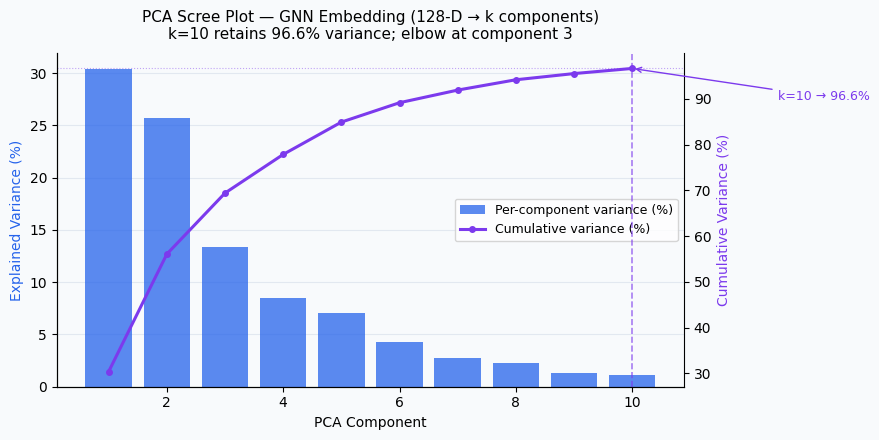}
\caption{PCA Scree Plot of 128-D GNN embeddings. Ten components
retain 96.6\% of variance; the sharp elbow at component~3 confirms
an inherently low-dimensional O2I propagation
manifold.}\label{fig:pca_scree}
\end{figure}

However, eigen-analysis revealed that this embedding space occupies
a strict low-dimensional manifold: the top principal components
consistently dominated the variance whether the hidden dimension was
set to 128 or 256, and forcing the XGBoost regressor to search the
raw 128-dimensional space caused rapid overfitting and early
stopping. Compressing the embeddings to a 10-dimensional subspace
($k=10$) retains 96.6\% of the total embedding variance
(Fig.~\ref{fig:pca_scree}). The elbow in per-component variance
drops sharply after the third component, indicating that the first
three carry the dominant spatial signal, components 4 through 10 add
localized refinement, and components 11 through 128 function largely
as high-dimensional topological noise.

By compressing the data to this $k=10$ subspace, we stripped the
high-dimensional noise, allowing the tree-based models to converge
efficiently on the true physical propagation axes. Notably,
post-PCA feature importance analysis revealed that the original raw
spatial coordinates contributed negligibly ($\approx$0\%) to the XGBoost splits. This
confirms that the GNN successfully learned a comprehensive
representation that entirely subsumed the raw geographic coordinates.

\subsection{3D Spatial Visualization and Coverage Mapping}

Beyond statistical metrics, the practical utility of the proposed
framework lies in its ability to generate high-resolution,
volumetric coverage maps. We developed a web-based visualization
tool utilizing Inverse Distance Weighting (IDW, $p=2$) interpolation
and D3.js contour generation to render continuous signal heatmaps
across all 10 building floors.

\begin{figure}[ht!]
\centering
\includegraphics[width=0.7\textwidth]{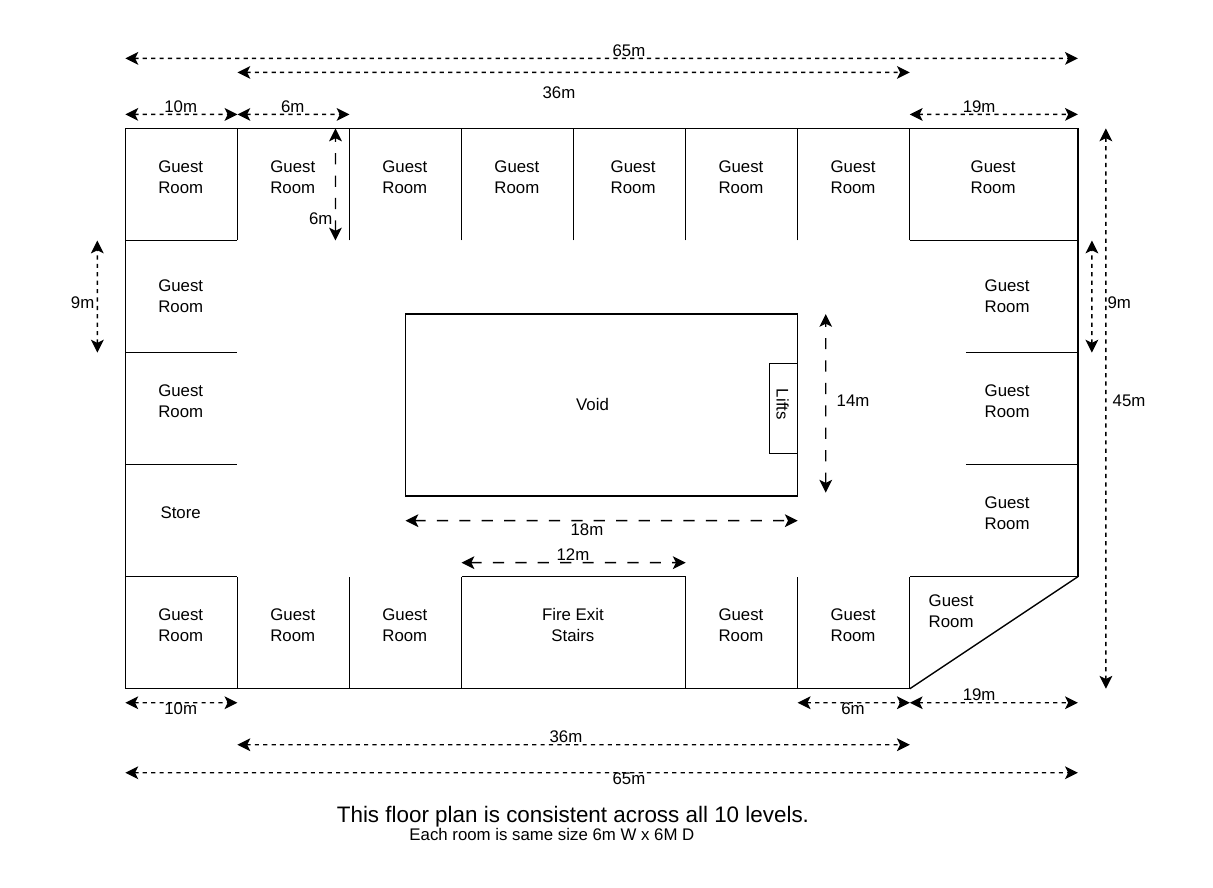}
\caption{Annotated building floorplan defining the physical
dimensions and corridor bounding box. This geometry is structurally
consistent across all 10 levels and acts as the spatial constraint
for the RF visualization engine.}\label{fig:floorplan}
\end{figure}

To project the inferred topological embeddings back into a
physically interpretable space, the building's architectural
floorplan was mapped at a 15~pixel/meter conversion ratio over a
65~m~$\times$~45~m footprint, with structural room depths taken from
architectural documentation and a 2-meter inset applied so that
signal data points do not render flush against structural corridor
walls. This layout is structurally consistent across all 10 levels
(Fig.~\ref{fig:floorplan}) and defines the physical corridor
bounding box that constrains the IDW signal interpolation,
preventing signal contours from bleeding through solid concrete
cores.

To provide actionable intelligence for network planners, the
predicted continuous signal values generated within these boundaries
are classified into three operational bands (Poor, Fair, Good)
following the qualitative signal-quality ranges commonly adopted in
network planning practice (Table~\ref{tab:signal_thresholds}).

\begin{table}[ht!]
\caption{Signal quality threshold classifications for contour
mapping.}\label{tab:signal_thresholds}
\centering
\begin{tabular}{|l|c|c|c|c|}
\hline
\textbf{Parameter} & \textbf{Unit}
  & \textbf{Poor (Red)} & \textbf{Fair (Yellow)}
  & \textbf{Good (Green)} \\
\hline
RSSI  & dBm   & $-110$ to $-100$ & $-100$ to $-70$ & $-70$ to $-10$ \\
RSRP  & dBm   & $-140$ to $-100$ & $-100$ to $-70$ & $-70$ to $-40$ \\
RSRQ  & dB    & $-30$ to $-24$   & $-24$ to $-9$   & $-9$ to $0$   \\
SNR   & dB    & $-40$ to $0$     & $0$ to $8$      & $8$ to $30$   \\
CQI   & Index & $0$ to $5$       & $5$ to $10$     & $10$ to $15$  \\
\hline
\end{tabular}
\end{table}

\begin{figure}[ht!]
\centering
\includegraphics[width=0.55\textwidth]{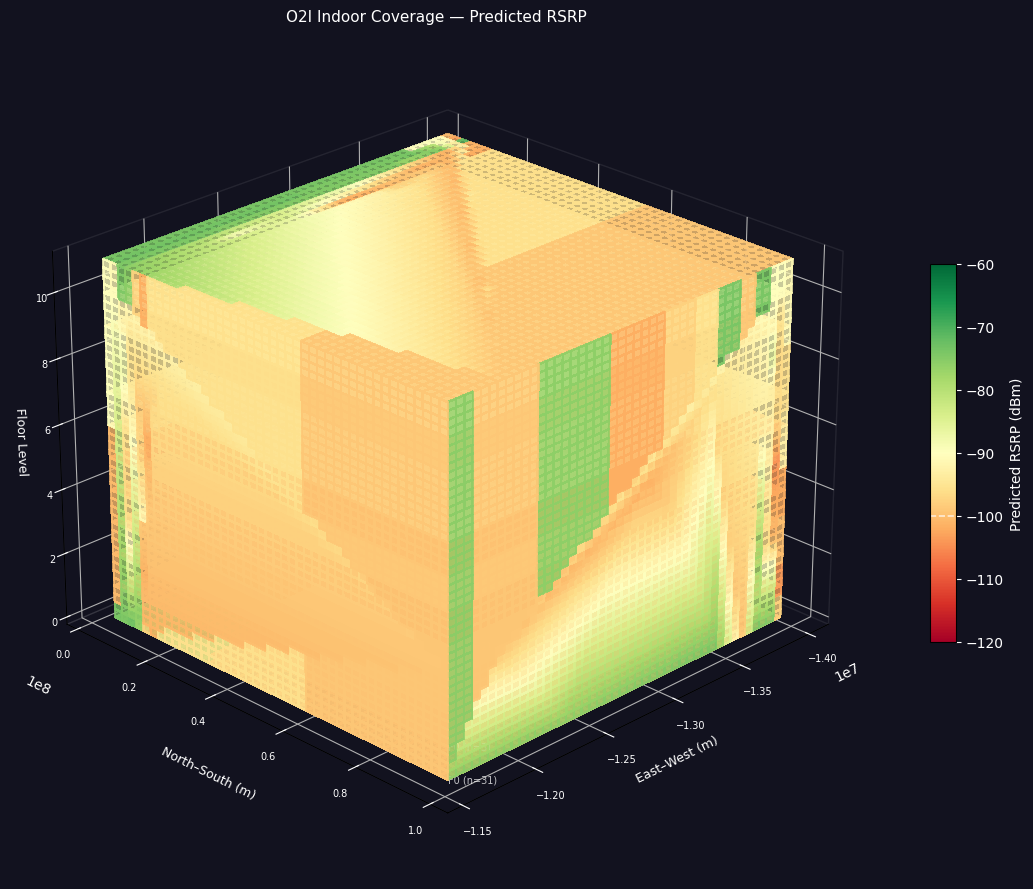}
\caption{3D volumetric visualization of the inferred In-Building
Coverage (IBC). The model stacks the 2D IDW heatmaps across
10 floors, revealing spatial attenuation patterns and signal
degradation along the vertical axis of the building
manifold.}\label{fig:3d_volumetric}
\end{figure}

Applying these thresholds to the IDW interpolated data, the
resulting 3D volumetric visualization (Fig.~\ref{fig:3d_volumetric})
successfully interpolates the sparse outdoor UAV boundary
measurements across all building levels. By rendering the predicted
signal attenuation dynamically across stacked floor textures, the
framework visually identifies multi-story structural dead zones
without requiring physical indoor entry.

\section{Conclusion}
This paper presented Graph-XGB, a seven-phase hybrid inference
framework for reconstructing the three-dimensional indoor RF
environment of multi-story structures from sparse exterior UAV
boundary measurements alone, with no interior RF value entering the
model as an input feature. The central architectural insight is the
deliberate decoupling of non-Euclidean spatial topology reasoning,
performed by a modified SpiderGNN encoder operating on a physically
grounded floor-aware graph, from non-linear signal regression
executed by independent, target-specific XGBoost ensembles, a
separation that directly mitigates the optimization conflict
afflicting end-to-end GNN approaches when predicting signals
governed by divergent physics. Evaluated on a real-world 5G NR
multi-modal dataset drawn from a single 10-floor building,
Graph-XGB achieved $R^2$ of 0.759 (RSSI) and 0.770 (RSRP),
improvements of 35.5\% and 64.9\% over SimpleGNN and of 71\% and
over 100\% over the end-to-end SpiderGNN predecessor, while the PCA
eigen-analysis confirmed an inherently low-dimensional O2I manifold
(10 components, 96.6\% variance). The framework further enables
high-resolution 3D volumetric heatmaps that localise dead zones
across all ten floors without further indoor walk-testing once
trained.

Two boundaries frame these results. Interference-dependent metrics
(RSRQ, SNR, CQI) remain bounded by a predictive ceiling
($R^2 \leq 0.27$) under purely spatial inference; this Signal-Type
Ceiling Hypothesis indicates that such volatile, load-dependent
metrics require temporal and network-state context beyond what
static spatial graphs can encode. Second, training and evaluation
draw from one building, one campaign and 5G NR measurements, so the
reported accuracy characterises reconstruction at unobserved
interior locations within an instrumented structure, and the 6G
sub-THz regime motivates rather than evidences this work. Future
work will therefore incorporate temporal network load and
neighbour-cell interference as dynamic GNN node features, extend to
multi-building generalisation via transfer learning across
heterogeneous morphologies, and explore differentiable alternatives
to PCA to jointly optimise embedding compression and downstream
decoding.

\begin{credits}
\subsubsection{\ackname}
The authors acknowledge the use of generative AI tools to improve
the readability and grammatical flow of this manuscript. The authors
have reviewed and edited the content and take full responsibility
for the final text.

\subsubsection{\discintname}
The authors declare that they have no known competing financial
interests or personal relationships that could have appeared to
influence the work reported in this paper.
\end{credits}

\bibliographystyle{unsrt}
\bibliography{mybibliography}

\begin{thebibliography}{10}

\bibitem{5gamericas2024neutral}
{5G Americas}.
\newblock Neutral host opportunities for {5G} \& beyond.
\newblock Technical report, 5G Americas, 2024.

\bibitem{ericsson2025mobility}
{Ericsson}.
\newblock Ericsson mobility report.
\newblock Technical report, Ericsson, June 2025.

\bibitem{xing2021sub}
Y~Xing and TS~Rappaport.
\newblock Sub-terahertz and mmwave penetration loss measurements for indoor
  environments.
\newblock {\em arXiv preprint arXiv:2103.02745}, 2021.

\bibitem{infovista2025tems}
Infovista.
\newblock Indoor and drone-based network testing and troubleshooting {TEMS}
  pocket v26.0.
\newblock \url{https://www.infovista.com}, 2025.
\newblock Accessed: 2025.

\bibitem{ficili2025sensors}
I~Ficili, M~Giacobbe, G~Tricomi, and A~Puliafito.
\newblock From sensors to data intelligence: Leveraging {IoT}, cloud, and edge
  computing with {AI}.
\newblock {\em Sensors}, 25(6):1763, 2025.

\bibitem{tarhuni2023machine}
N~Tarhuni, I~Al~Saadi, HM~Asif, M~Mesbah, O~Eldirdiry, and A~Hossen.
\newblock Machine-learning-based ground-level mobile network coverage
  prediction using {UAV} measurements.
\newblock {\em Journal of Sensor and Actuator Networks}, 12(3):44, 2023.

\bibitem{wang2014cellular}
C-X Wang, F~Haider, X~Gao, X-H You, Y~Yang, D~Yuan, et~al.
\newblock Cellular architecture and key technologies for {5G} wireless
  communication networks.
\newblock {\em IEEE Communications Magazine}, 52(2):122--130, 2014.

\bibitem{shakya2022dense}
D~Shakya, D~Chizhik, J~Du, RA~Valenzuela, and TS~Rappaport.
\newblock Dense urban outdoor-indoor coverage from 3.5 to 28 {GHz}.
\newblock In {\em ICC 2022-IEEE International Conference on Communications},
  pages 932--937. IEEE, 2022.

\bibitem{karner2006indoor}
W~Karner, A~Paier, and M~Rupp.
\newblock Indoor coverage prediction and optimization for {UMTS} macro cells.
\newblock In {\em 2006 3rd International Symposium on Wireless Communication
  Systems}, pages 625--630. IEEE, 2006.

\bibitem{mostafa2025survey}
Sherif Mostafa, Khaled~A Harras, and Moustafa Youssef.
\newblock A survey of indoor localization systems for multi-floor environments.
\newblock {\em IEEE Access}, 13, 2025.

\bibitem{ng2022electromagnetic}
JPS Ng, YL~Sum, BH~Soong, M~Maier, and PJ~Monteiro.
\newblock Electromagnetic wave propagation through composite building materials
  in urban environments at mid-band {5G} frequencies.
\newblock {\em IET Microwaves, Antennas \& Propagation}, 16(10):627--638, 2022.

\bibitem{ayeni2022correlation}
AA~Ayeni, AA~Yusuf, and SO~Onidare.
\newblock Correlation between the electromagnetic properties of building
  materials and wireless signal penetration loss.
\newblock {\em IEEE Transactions on Antennas and Propagation},
  70(12):12040--12048, 2022.

\bibitem{pavlik2024influence}
M~Pavl{\'\i}k, M~Bere{\v{s}}, and L~Be{\v{n}}a.
\newblock The influence of various commonly used building materials on the
  shielding effectiveness, reflection and absorption of the electromagnetic
  wave.
\newblock {\em Applied Sciences}, 14(6):2521, 2024.

\bibitem{asp2012radio}
A~Asp, Y~Sydorov, M~Valkama, and J~Niemel{\"a}.
\newblock Radio signal propagation and attenuation measurements for modern
  residential buildings.
\newblock In {\em 2012 IEEE Globecom Workshops}, pages 580--584. IEEE, 2012.

\bibitem{fuschini2022multiband}
Franco Fuschini, Marina Barbiroli, Enrico~Maria Vitucci, and Vittorio
  Degli-Esposti.
\newblock Multi-band outdoor-to-indoor propagation measurements using a drone.
\newblock In {\em 2022 16th European Conference on Antennas and Propagation
  (EuCAP)}, pages 1--5. IEEE, 2022.

\bibitem{ghanem2012reducing}
K~Ghanem, H~Alradwan, A~Motermawy, and A~Ahmad.
\newblock Reducing ping-pong handover effects in intra {EUTRA} networks.
\newblock In {\em 2012 8th International Symposium on Communication Systems,
  Networks \& Digital Signal Processing (CSNDSP)}, pages 1--5. IEEE, 2012.

\bibitem{ma2024reconfigurable}
C~Ma, X~Yang, J~Wang, G~Yang, W~Zhang, and S~Ma.
\newblock Reconfigurable distributed antennas and reflecting surface: A new
  architecture for wireless communications.
\newblock {\em IEEE Transactions on Communications}, 2024.

\bibitem{jafari2015small}
AH~Jafari, D~L{\'o}pez-P{\'e}rez, H~Song, H~Claussen, L~Ho, and J~Zhang.
\newblock Small cell backhaul: challenges and prospective solutions.
\newblock {\em EURASIP Journal on Wireless Communications and Networking},
  2015:1--18, 2015.

\bibitem{paulo2021study}
RR~Paulo, FJ~Velez, and B~Khan.
\newblock Study of indoor small cell deployments.
\newblock {\em Journal of Mobile Multimedia}, pages 329--344, 2021.

\bibitem{fotouhi2019survey}
A~Fotouhi, H~Qiang, M~Ding, M~Hassan, LG~Giordano, et~al.
\newblock Survey on {UAV} cellular communications: Practical aspects,
  standardization advancements, regulation, and security challenges.
\newblock {\em IEEE Communications Surveys \& Tutorials}, 21(4):3417--3442,
  2019.

\bibitem{martian2024direction}
A~Martian, C~Paleacu, I-M Marcu, and C~Vladeanu.
\newblock Direction-finding for unmanned aerial vehicles using radio frequency
  methods.
\newblock {\em Measurement}, 235:114883, 2024.

\bibitem{rocha2023ziwi}
L~Rocha, S~Ferreira, KCT Vivaldini, J~Ara{\'u}jo, and I~Batalha.
\newblock Ziwi: indoor and outdoor planning network-framework to collection,
  modeling and network structure based on computational optimization and
  measurements.
\newblock {\em Soft Computing}, 27(10):6761--6781, 2023.

\bibitem{amano2025referencefree}
Tatsuya Amano, Hirozumi Yamaguchi, and Teruo Higashino.
\newblock Reference-free 3d wifi ap localization by outdoor-to-indoor bridging.
\newblock {\em IEEE Open Journal of the Computer Society}, 2025.

\bibitem{kurner2002prediction}
T~Kurner and A~Meier.
\newblock Prediction of outdoor and outdoor-to-indoor coverage in urban areas
  at 1.8 {GHz}.
\newblock {\em IEEE Journal on Selected Areas in Communications},
  20(3):496--506, 2002.

\bibitem{maduraga2021comparison}
M~Maduraga and R~Abeysekara.
\newblock Comparison of supervised learning-based indoor localization
  techniques for smart building applications.
\newblock In {\em 2021 International Research Conference on Smart Computing and
  Systems Engineering (SCSE)}, volume~4, pages 145--148. IEEE, 2021.

\bibitem{raj2021indoor}
N~Raj.
\newblock Indoor {RSSI} prediction using machine learning for wireless
  networks.
\newblock In {\em 2021 International Conference on Communication Systems \&
  Networks (COMSNETS)}, pages 372--374. IEEE, 2021.

\bibitem{konak2024wireless}
A~Konak, S~Delattre, and M~Bartolacci.
\newblock Wireless signal prediction using deep learning models for {WiFi}
  positioning and security concerns.
\newblock {\em Wireless Personal Communications}, 2024.

\bibitem{qiu2023irdm}
Kehai Qiu, Stefanos Bakirtzis, Ian Wassell, Hui Song, Kan Lin, and Jie Zhang.
\newblock Irdm: A generative diffusion model for indoor radio map
  interpolation.
\newblock In {\em 2023 IEEE Global Communications Conference (GLOBECOM)}, pages
  1--6. IEEE, 2023.

\bibitem{wang2025iradiodiff}
Xiucheng Wang, Tingwei Yuan, Yang Cao, Nan Cheng, Ruijin Sun, and Weihua
  Zhuang.
\newblock iradiodiff: Physics-informed diffusion model for indoor radio map
  construction and localization.
\newblock {\em arXiv preprint arXiv:2511.20015}, 2025.

\bibitem{wang2026radiodiff}
Xiucheng Wang, Qiming Zhang, Nan Cheng, Ruijin Sun, Zan Li, Shuguang Cui, and
  Xuemin Shen.
\newblock Radiodiff-k2: Helmholtz equation informed generative diffusion model
  for multi-path aware radio map construction.
\newblock {\em IEEE Journal on Selected Areas in Communications}, 44, 2026.

\bibitem{zhang2023domain}
M~Zhang, Z~Fan, R~Shibasaki, and X~Song.
\newblock Domain adversarial graph convolutional network based on {RSSI} and
  crowdsensing for indoor localization.
\newblock {\em IEEE Internet of Things Journal}, 10(15):13662--13672, 2023.

\bibitem{lezama2021indoor}
F~Lezama, GG~Gonz{\'a}lez, F~Larroca, and G~Capdehourat.
\newblock Indoor localization using graph neural networks.
\newblock In {\em 2021 IEEE URUCON}, pages 51--54. IEEE, 2021.

\bibitem{lezama2023application}
F~Lezama, F~Larroca, and G~Capdehourat.
\newblock On the application of graph neural networks for indoor positioning
  systems.
\newblock In {\em Machine Learning for Indoor Localization and Navigation},
  pages 239--256. Springer, 2023.

\bibitem{vishwakarma2023indoorgnn}
R~Vishwakarma, RB~Joshi, and S~Mishra.
\newblock Indoorgnn: A graph neural network based approach for indoor
  localization using {WiFi} {RSSI}.
\newblock In {\em International Conference on Big Data Analytics}, pages
  150--165. Springer, 2023.

\bibitem{bayraktar2024indoor}
Mert Bayraktar and Alper Ozcan.
\newblock Indoor localization with gravnetconv and dynamic graphs.
\newblock In {\em 2024 9th International Conference on Computer Science and
  Engineering (UBMK)}, pages 1--5. IEEE, 2024.

\bibitem{zhang2026graphkanloc}
Hao Zhang, Yunfan Jiang, Xin Deng, and Kai Liu.
\newblock Graphkanloc: Physics-informed heterogeneous graph attention networks
  with kan regression for indoor localization.
\newblock {\em IEEE Transactions on Mobile Computing}, 2026.

\bibitem{gao2025spidergnn}
Xingyi Gao, David~Hason Rudd, Ziang Li, Yuming Guo, Huan Huo, and Guandong Xu.
\newblock Spidergnn: Spatial-aware predictive inference with dynamic edge
  reasoning for in-building 5g signal estimation.
\newblock In {\em International Conference on Behavioural and Social
  Computing}, pages 123--140. Springer, 2025.

\bibitem{rudd2025location}
David~Hason Rudd, Cesar Sanin, Koh~Ming En, Xingyi Gao, Md~Rafiqul Islam,
  Mehedi Hasan, Xianzhi Wang, Angela Huo, and Guandong Xu.
\newblock Location-aware deep neural network for predicting indoor 5g rssi and
  cqi using drone-based external rf sensing.
\newblock {\em Procedia Computer Science}, 270:4765--4775, 2025.

\end{thebibliography}

\end{document}